\documentclass{ecai} 

\usepackage{latexsym}
\usepackage{amssymb}
\usepackage{amsmath}
\usepackage{amsthm}
\usepackage{booktabs}
\usepackage{enumitem}
\usepackage{graphicx}
\usepackage{color}
\usepackage{algorithm}
\usepackage{algorithmicx}
\usepackage{algpseudocode}
\usepackage[table,xcdraw]{xcolor}

\newcommand{\BibTeX}{B\kern-.05em{\sc i\kern-.025em b}\kern-.08em\TeX}

\begin{document}


\begin{frontmatter}


\paperid{123} 


\title{Monetizing Currency Pair Sentiments \\ through LLM Explainability}


\author[A]{\fnms{Lior}~\snm{Limonad}\thanks{Corresponding Author. Email: liorli@il.ibm.com}}
\author[B]{\fnms{Fabiana}~\snm{Fournier}}
\author[C]{\fnms{Juan Manuel}~\snm{Vera Díaz}} 
\author[D]{\fnms{Inna}~\snm{Skarbovsky}}
\author[E]{\fnms{Shlomit}~\snm{Gur}}
\author[G]{\fnms{Raquel}~\snm{Lazcano}}

\address[A,B,D,E]{IBM Research, Israel}
\address[C,G]{Atos IT Solutions and Services Iberia, Madrid, Spain}


\begin{abstract}
Large language models (LLMs) play a vital role in almost every domain in today’s organizations. In the context of this work, we highlight the use of LLMs for sentiment analysis (SA) and explainability. Specifically, we contribute a novel technique to leverage LLMs as a post-hoc model-independent tool for the explainability of SA. We applied our technique in the financial domain for currency-pair price predictions using open news feed data merged with market prices. Our application shows that the developed technique is not only a viable alternative to using conventional eXplainable AI but can also be fed back to enrich the input to the machine learning (ML) model to better predict future currency-pair values. 
We envision our results could be generalized to employing explainability as a conventional enrichment for ML input for better ML predictions in general.
\end{abstract}

\end{frontmatter}


\newif\ifshowcomments
\showcommentstrue
\ifshowcomments
\newcommand{\mynote}[2]{\fbox{\bfseries\sffamily\scriptsize{#1}}
{\small$\blacktriangleright$\textsf{#2}$\blacktriangleleft$}}
\else
\newcommand{\mynote}[2]{}
\fi
\newcommand{\inna}[1]{\textcolor{red}{\mynote{Inna}{#1}}}
\newcommand{\lior}[1]{\textcolor{blue}{\mynote{LIOR}{#1}}}
\newcommand{\fabiana}[1]{\textcolor{red}{\mynote{Fabiana}{#1}}}
%

\section{Introduction}

Explainability is the foundation for the adoption and trust of humans in AI-based systems. Explanations are the vehicle via which users can understand and act upon the various situations that evolve during their interaction with the system. 
Although some Machine Learning (ML) models are inherently explainable (e.g., decision trees and linear models) and their internal logic predictions are interpretable (i.e., can be easily understood by humans), more complex models (e.g., ensemble models and deep learning models) require external explanation frameworks, namely eXplainable AI (XAI), to be human-understandable~\cite{Guidotti2018,Dumas2023}. 
Probably most prominent recently is the employment of Generative AI\footnote{\url{www.gartner.com/en/information-technology/glossary/generative-ai}}, and particularly its applicability to textual artifacts in the form of Large Language Models (LLMs). This presents a promising instrumentation to enable out-of-the-box model-independent explanations that are easy to interpret by humans.

Gartner's report~\cite{Gartner2023} highlights the widespread adoption of LLMs across industries 
for uses including text summarization, question-answering, and document translation. LLMs can also be augmented by additional capabilities to create more powerful systems and feature a growing ecosystem of tools. Among these, we foresee the benefit of leveraging LLMs not only as a means for the automation of explanations in AI-based systems but also 
as an enrichment mechanism that could be fed back as input for the AI itself to improve its performance. Our results in this work pave the way to conducting a more rigorous study to support our claim.

As one possible application of XAI in the financial sector, we have developed an application that leverages ML for \textit{\emph\ Sentiment Analysis (SA)} on news feeds about selected currency pairs. 
SA has become essential for grasping market dynamics and forecasting trends in the financial sector. This technique's ability to gauge the overall mood of the market offers critical insights, thereby facilitating more informed and strategic decision-making.
For each, a collection of related news articles is gathered over time. Our contribution is in showing not only that the series of sentiments extracted from such feeds can be leveraged to predict currency-pair rate changes but also to populate information that can be used to enrich and thus improve the accuracy of such predictions. 

\begin{figure*}
    \centering
    \includegraphics[width=0.9\linewidth]{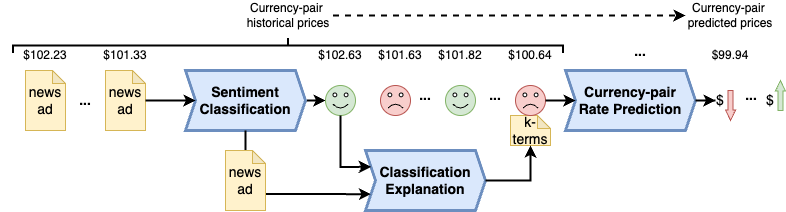}
    \caption{Overall approach}
    \label{fig:overall-process}
\end{figure*}

The overall high-level approach is depicted in Figure~\ref{fig:overall-process}. 
Our ultimate goal is to explore the potential to predict financial projections about currency-pair rates based on the underlying sentiment sequences. Thus, given an input of a series of daily currency-pair rates accompanied by corresponding news ads, as a first step, we employ sentiment classification over these news ads to derive their corresponding sentiment. Our idea is that accompanying currency-pair prices with their associated sentiments could improve price predictions. As a second step, we developed a novel algorithm that employs LLMs to extract the set of terms explaining each sentiment classification. As a last step, we blend these terms with the sentiments and the original price stream as an enriched input for the rate prediction model.

\section{Background}

Recent advancements in the ML field have increased the complexity of ML models, often at the expense of their interpretability, leading to ``black box'' ML models and hindering their full adoption. Therefore, we have also seen an increase in the need to explain ML models and their predictions, fuelled in part by legislation, but also by incentives from the user’s or stakeholder’s point of view (e.g., justify the ML model and gain domain insight), and from the developer’s point of view (e.g., evaluate and improve the ML model). 

\textit{\emph\ eXplainable AI (XAI)} is a research field that aims to make AI systems results more understandable to humans~\cite{Adadi2018}. Many XAI frameworks are predominately developed for post-hoc (i.e., after model training) interpretations of ML models. 
In contemporary XAI techniques (e.g., LIME~\cite{Ribeiro2016} or SHAP~\cite{Lundberg2017}), the ML model serves as a surrogate model typically trained using historical data. The predicted value for a single instance serves as input for the XAI explainer to produce an explanation.

Leveraging LLMs for the sake of explainability eliminates the burden of training such explainability ML models. 
\textit{\emph\ LLMs} are deep-learning models trained on vast amounts of text data to perform various natural language processing tasks. One of their strengths is their ability to perform few-shot and zero-shot learning with prompt-based learning~\cite{LLMprompt2023}. With the emergence of LLMs, the practice of `prompt engineering' has been explored recently as a new field of research that focuses on designing, refining, and implementing different instructions to guide the output of LLMs in various tasks to achieve a more concise and effective utilization of the LLMs, adapting it to the corresponding skills of the users and the context of the different tasks such as analysis or reasoning~\cite{Mesko2023}. This may include different techniques such as one- or few-shot samples, the use of quotes to annotate different parts of a prompt, employing methodologies 
such as ``Chain of Thought'' (CoT), and configuration of LLM-specific settings like temperature and top-p sampling to control the variability of the output.

In this work, we employ SA for a concrete financial application of predicting currency-pair prices.
\textit{SA} is the computational study of opinions, sentiments, and emotions expressed in text~\cite{Bing2012}. 
As a field of Natural Language Processing (NLP), a variety of techniques may be employed nowadays to enable SA, ranging from basic rule-based methods to ML and deep learning-based approaches, that typically involve model training based on corpora of textual narratives that are labeled by sentiment data. In this work, we present an LLM-based scheme for explaining SA that is agnostic to the SA technique employed.


\section{LLM for SA explainability}
\label{sec:llm-4-sax}

We developed a technique for the use of an LLM for model-agnostic explainability, given as an input a narrative and its sentiments as determined by some SA model (e.g., VADER or BERT). 
Our technique 
identifies the set of keywords or terms in the narrative that support (i.e., explain) the sentiment according to the inference model.

Our approach was developed as a post-hoc technique to enable the identification of the `K' most relevant set of keywords or terms (i.e., the k-sufficient set) in the input narrative that is sufficient to influence the prediction of the ML model, regardless of which specific ML technique is used for the prediction of the sentiment. Concretely, a k-sufficient set is a subset of words from the original narrative that when provided as an input to the ML model retains the same sentiment output as the sentiment that was originally generated for the entire narrative. Thus, given some model $M$ and an input narrative text $T\ =\ w_1,...,w_n$, we can determine its sentiment as $M(T)$. Respectively, a k-sufficient set in this case is a subset of $k$ terms  $s_k=w_1,...,w_k\subseteq T$, where $\left|s_k\right|=k$, conforming to:

\begin{enumerate}
    \item $M\left(s_k\right)=M\left(T\right)$, i.e., the subset of terms retains the same sentiment.
    \item $\bigl[$optional$\bigr]$ There is no subset $s\prime=w_1...w_z\subset$ $s_k$ where $M(s\prime)=M(T)$
\end{enumerate}

The set $s_k$ is deemed sufficient in the sense that its inclusion in text $T$ guarantees the sentiment classification of $T$ equals the original sentiment classification as deemed by the ML model (i.e., $M(T)$). We acknowledge that there could be several sets in $T$ conforming to the above conditions. Our current implementation is indifferent among such subsets.

Step 1 in our method accounts for explanation sufficiency corresponding to the input request for K terms. That is, such a set contains all needed terms to ensure it preserves consistency with the sentiment classification of the original narrative. This does not assure that such a set contains the fewest terms necessary to retain the original sentiment classification (i.e., a minimal set). If such a requirement is also deemed necessary, we include step two as an optional extension to fulfill such a requirement. Without concerns of performance, a straightforward realization will need to add to the given implementation also an exhaustive scan of all $\left|k-1\right|$ subsets $s\prime$ to ensure that for any such subset $M\left(s\prime\right)\neq M(T)$. Another possible realization may directly leverage on the power of the LLM to identify such a minimal subset directly.

\begin{figure*}[ht]
    \centering
    \includegraphics[width=0.9\linewidth]{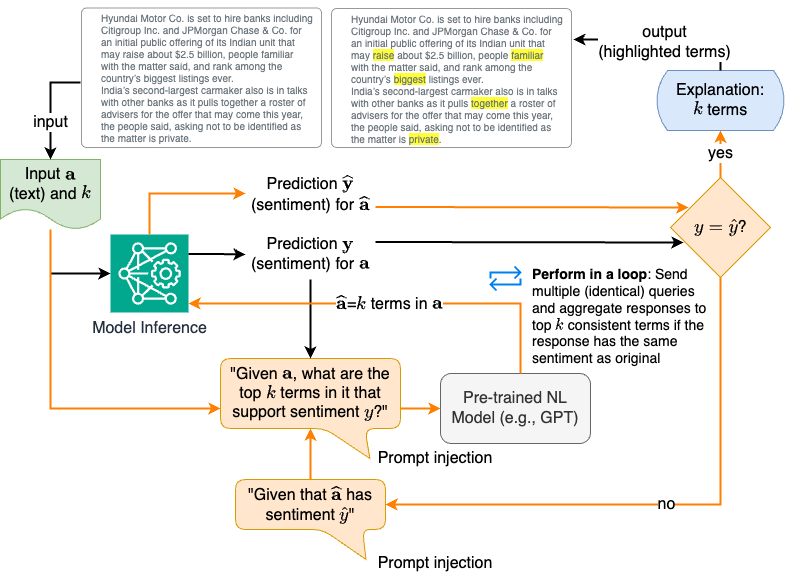}
    \caption{Explanation of SA with k-sufficient terms}
    \label{fig:SA-explanation}
\end{figure*}

A more advanced approach could also exploit the computation of weights that signify the relative importance of each word in affecting the result of the ML model. In such a case, our algorithm should be enhanced with a selection of a subset that relies on the value of such weights (e.g., the one with the highest weight average). To elicit a k-sufficient set of keywords from a given narrative, we follow a zero-shot prompting approach. 
The overall flow of our approach to classification explanation is depicted in Figure~\ref{fig:SA-explanation}.

Algorithm~\ref{alg:llm-sentiment-explain} formally elaborates on the realization of the classification explanation, given $M$ as the model for sentiment elicitation, $X$ as a textual narrative, $L$ is the employed LLM, $K$ as the number of keywords, and $maxAttempts$ as a threshold limit to the number of iteration attempts. An implementation of the algorithm is available here: \url{https://github.com/IBM/SAX/AlFin}.

\algdef{SE}[DOWHILE]{Do}{doWhile}{\algorithmicdo}[1]{\algorithmicwhile\ #1}%
\begin{algorithm}[!ht]
\caption{\textsc{LLMsentimentXplain}($M$, $X$, $L$, $K$, $maxAttempts$)}\label{alg:llm-sentiment-explain}
\begin{algorithmic}[1]
\State $result \gets \text{``no explanation''}$
\State $s \gets M.\text{getSentiment}(X)$
\State $attempts \gets 0$
\State $prompt \gets \text{``given  text:''}+X+\text{``what are the top''} + K +$
\Statex $\text{``terms that support its sentiment classification as:''}+s$
\Do
\State $attempts \gets attempts+1$
\State $W \gets L.\text{apply}(prompt)$
\State $s\prime \gets M.\text{getSentiment}(W)$
\If{$s \neq s\prime$} 
    \State $prompt \gets prompt+\text{``the sentiment classification of:''}+W+\text{``is''}+s\prime$
\EndIf 
\doWhile{$(s \neq s\prime) \And (attempts < maxAttempts)$} 
\If{$s = s\prime$}
    \State $result \gets W$
\EndIf
\State \Return $result$
\end{algorithmic}
\end{algorithm}




\section{Enriched feeds to predict currency-pair prices}


Building on prior work~\cite{Fatouros2023}, our first envisioned hypothesis is that enriching an input feed of price closing rates with sentiment information that corresponds to a series of news ads about a given currency pair may promote not only a better prediction of future trends (i.e., increase or decrease in price) but also to the improved predictive accuracy of future closing prices. By enrichment of the input, we relate to extending the input with additional term embeddings as further elaborated below. 

Going beyond the first hypothesis, we further envision the potential monetary benefit to be exploited from explanatory information about sentiment classification. Leveraging the technique presented in section~\ref{sec:llm-4-sax} for associating each sentiment with a set of k-sufficient terms for a given news ad, our next step is to use this technique to enrich a series of news-ad-derived sentiments with their corresponding highlighted keywords. We envision that such enriched traces may serve as a valuable input for the prediction of future currency indicators, correlating with price movements such as currency-pair closure prices (i.e., as in~\cite{Fatouros2023}) and that complementing such sentiment traces with the k-sufficient keywords (along each sentiment label) may help to improve the accuracy of such predictions. With the technique presented, we also foresee the viability of deriving such information automatically with the use of recently developed LLMs.

\noindent
Hence we hypothesize that:
\begin{itemize}
\vspace{-0.5em}
    \item \textbf{H1:} News-ad-derived sentiment information can be used to improve the accuracy of predictions about currency-pair future rate behaviors.  

    \item \textbf{H2:} Sentiment feeds enriched with explanatory information will achieve better predictive accuracy than non-enriched sentiment feeds.
\vspace{-0.5em}
\end{itemize}

To evaluate the hypotheses, we used an open sentiment labeled dataset from~\cite{Fatouros2023}. 
This data consists of 2291 news ads about five different currency pairs collected between January and May 2023. 
This dataset was combined with daily currency-pair historical prices for the same period as summarized in Table~\ref{tab:close-prices}, which was downloaded from the Yahoo Finance website. Note that the \texttt{USDJPY} price is about two orders of magnitude greater than all other currency pair prices.

\begin{table*}[htbp]
\caption{Currency-pair closing prices during the 3 months period}
\centering
\resizebox{0.8\textwidth}{!}{%
\begin{tabular}{|l|l|l|l|l|l|}
\hline
\textbf{Currency-pair} & \textbf{AUDUSD} & \textbf{EURCHF} & \textbf{EURUSD} & \textbf{GBPUSD} & \textbf{USDJPY} \\ \hline
\textbf{Max close price} & 0.7145 & 1.0069 & 1.1068 & 1.2630 & 140.8710 \\ \hline
\textbf{Min close price} & 0.6498 & 0.9684 & 1.0522 & 1.1827 & 128.0260 \\ \hline
\textbf{Average Close Price} & 0.6773 & 0.9876 & 1.0806 & 1.2278 & 133.5069 \\ \hline
\textbf{Close Price STDev} & 0.0150 & 0.0090 & 0.0141 & 0.0198 & 2.9260 \\ \hline
\end{tabular}%
}
\label{tab:close-prices}
\end{table*}

Utilizing the dataset, a Long Short-Term Memory (LSTM) model~\cite{Hochreiter1997} was structured with a  32-unit layer followed by a neuron that regresses the currency price. An LSTM model is a recurrent neural network architecture well-suited for sequential data. The model was trained as a baseline to predict the next day's closing price from the past 5 days. To test the first hypothesis, we extended this input with three sentiment-derived features, denoting the percentage of news per day related to positive, neutral, and negative sentiment.

To test the second hypothesis, we first employed three different LLMs for the elicitation of k-sufficient keywords: IBM-granite-13b-chat, GPT-4.0, and GPT-3.5-turbo, the former model provided by IBM watsonx.ai\footnote{https://www.ibm.com/products/watsonx-ai/foundation-models} and the latter two by openAI\footnote{https://platform.openai.com/}. The use of GPT-3.5-turbo was also manually inspected to ensure the output was properly formed. Further to the elicitation of the keywords, we computed the embeddings of these keywords with an NNLM encoder~\cite{Bengio2003} and averaged it each day. This average was added as an additional input feature to the LSTM model. We repeated the process for each LLM model type separately.


\section{Results}

\begin{figure*}[ht]
    \centering
    \includegraphics[width=0.9\linewidth]{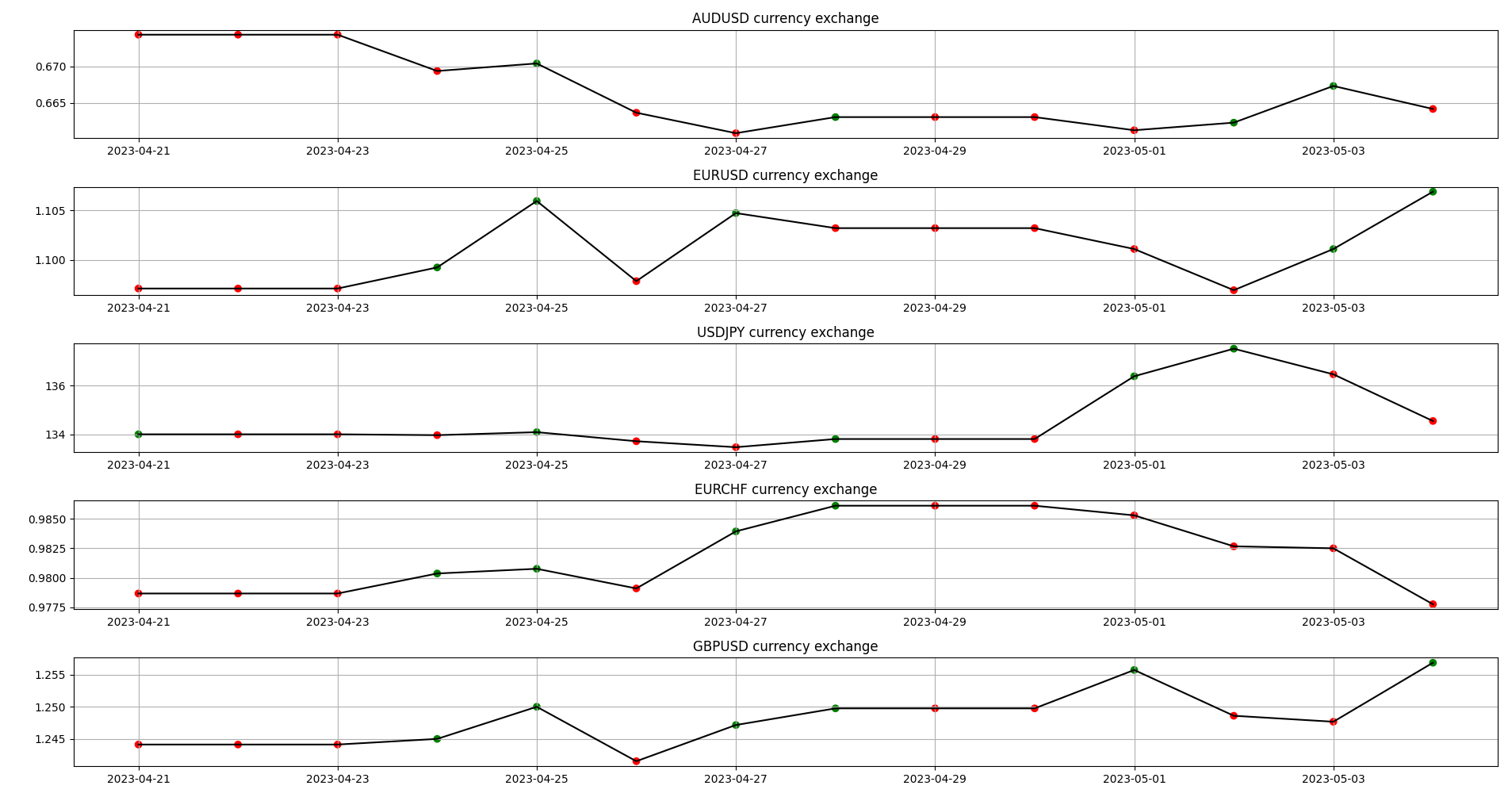}
    \caption{A test set example: 14 consecutive days for testing trend predictions. Red dots denote a drop and green dots denote a rise in currency-pair price with respect to the previous day.)}
    \label{fig:trend-prediction-test-set-example}
\end{figure*}

To evaluate future trend predictions the data consisted of 74 prediction points for each currency pair. Given days must be consecutive to maintain the temporal correlation, data was utilized in two separate instances: initially with a train-to-test split ratio of 44:30 days (i.e., $60\%/40\%$), and subsequently with a train-to-test split ratio of 60:14 days (i.e., $80\%/20\%$). An example of the latter case consisting of a test set of 14 consecutive days is shown in Figure~\ref{fig:trend-prediction-test-set-example}.
 
Similar to prior work, we assessed the quality of the predictions with an \texttt{accuracy} metric that captures the proportion of correct predictions out of the total number of predictions. Results for trend predictions are shown in Tables~\ref{tab:future-trends-30}~and~\ref{tab:future-trends-14}. The results in the tables are color-coded according to the range of each column's values, with higher accuracy represented by greener colors and lower accuracy indicated by yellower colors.

Corresponding to H1, on average the inclusion of the sentiment information on top of the underlying pricing information, as the input to the LSTM model did not yield any improvement in trend prediction accuracy. However, excluding the \texttt{EURCHF} currency pair, the embedding of the sentiment yielded either equal or improved accuracy in trend predictions.
With respect to H2, the inclusion of the explanation information was mostly effective for the GPT4.0 model, with all accuracy results except one (for \texttt{USDJPY} in Table~\ref{tab:future-trends-30}) being either the same or improved compared to the accuracy without the explanation information. For both remaining models, Granite and GPT-3.5 Curated, the inclusion of explanation information led to improved results on average, with both being mostly dominated by GPT4.0.

\begin{table*}[!ht]
\centering
\caption{Predicting future trends for currency pairs (accuracy metric), 44 consecutive days per currency for training and 30 consecutive days for testing}
\label{tab:future-trends-30}
\Large
\centering
\def\arraystretch{1.5}
\resizebox{\textwidth}{!}{%
\begin{tabular}{rrrrrrrr}
\multicolumn{1}{l}{{\color[HTML]{212121} }} & \multicolumn{1}{l}{{\color[HTML]{212121} }} & \multicolumn{1}{l}{{\color[HTML]{212121} \textbf{AUDUSD}}} & \multicolumn{1}{l}{{\color[HTML]{212121} \textbf{EURUSD}}} & \multicolumn{1}{l}{{\color[HTML]{212121} \textbf{USDJPY}}} & \multicolumn{1}{l}{{\color[HTML]{212121} \textbf{EURCHF}}} & \multicolumn{1}{l}{{\color[HTML]{212121} \textbf{GBPUSD}}} & \multicolumn{1}{l}{\textbf{Average}} \\
{\color[HTML]{212121} LSTM} & {\color[HTML]{212121} ACC} & \cellcolor[HTML]{CBDC81}{\color[HTML]{212121} 66,67\%} & \cellcolor[HTML]{E0E383}{\color[HTML]{212121} 60,00\%} & \cellcolor[HTML]{D6E082}{\color[HTML]{212121} 63,33\%} & \cellcolor[HTML]{ACD380}{\color[HTML]{212121} 76,67\%} & \cellcolor[HTML]{EBE583}{\color[HTML]{212121} 56,67\%} & \cellcolor[HTML]{D2DE82}{\color[HTML]{212121} 64,67\%} \\
{\color[HTML]{212121} LSTM+Sentiments} & {\color[HTML]{212121} ACC} & \cellcolor[HTML]{C1DA81}{\color[HTML]{212121} 70,00\%} & \cellcolor[HTML]{CCDD82}{\color[HTML]{212121} 66,66\%} & \cellcolor[HTML]{CBDC81}{\color[HTML]{212121} 66,67\%} & \cellcolor[HTML]{FED980}{\color[HTML]{212121} 43,35\%} & \cellcolor[HTML]{EBE583}{\color[HTML]{212121} 56,67\%} & \cellcolor[HTML]{DEE283}{\color[HTML]{212121} 60,67\%} \\
{\color[HTML]{212121} LSTM+Sentiments+Explanations (granite)} & {\color[HTML]{212121} ACC} & \cellcolor[HTML]{C1DA81}{\color[HTML]{212121} 70,00\%} & \cellcolor[HTML]{CBDC81}{\color[HTML]{212121} 66,67\%} & \cellcolor[HTML]{A2D07F}{\color[HTML]{212121} 80,00\%} & \cellcolor[HTML]{F5E984}{\color[HTML]{212121} 53,33\%} & \cellcolor[HTML]{CBDC81}{\color[HTML]{212121} 66,67\%} & \cellcolor[HTML]{C9DC81}{\color[HTML]{212121} 67,33\%} \\
{\color[HTML]{212121} LSTM+Sentiments+Explanations (gpt)} & {\color[HTML]{212121} ACC} & \cellcolor[HTML]{C1DA81}{\color[HTML]{212121} 70,00\%} & \cellcolor[HTML]{CBDC81}{\color[HTML]{212121} 66,67\%} & \cellcolor[HTML]{D6E082}{\color[HTML]{212121} 63,33\%} & \cellcolor[HTML]{ACD380}{\color[HTML]{212121} 76,67\%} & \cellcolor[HTML]{A2D07F}{\color[HTML]{212121} 80,00\%} & \cellcolor[HTML]{BDD881}{\color[HTML]{212121} 71,33\%} \\
{\color[HTML]{212121} LSTM+Sentiments+Explanations (gpt cur)} & {\color[HTML]{212121} ACC} & \cellcolor[HTML]{C1DA81}{\color[HTML]{212121} 70,00\%} & \cellcolor[HTML]{A2D07F}{\color[HTML]{212121} 80,00\%} & \cellcolor[HTML]{CBDC81}{\color[HTML]{212121} 66,67\%} & \cellcolor[HTML]{F5E984}{\color[HTML]{212121} 53,33\%} & \cellcolor[HTML]{CBDC81}{\color[HTML]{212121} 66,67\%} & \cellcolor[HTML]{C9DC81}{\color[HTML]{212121} 67,33\%}
\end{tabular}%
}
\end{table*}

\begin{table*}[!ht]
\centering
\caption{Predicting future trends for currency pairs (accuracy metric), 60 consecutive days per currency for training and 14 consecutive days for testing}
\label{tab:future-trends-14}
\Large
\centering
\def\arraystretch{1.5}
\resizebox{\textwidth}{!}{%
\begin{tabular}{rrrrrrrr}
\multicolumn{1}{l}{{\color[HTML]{212121} }} & \multicolumn{1}{l}{{\color[HTML]{212121} }} & \multicolumn{1}{l}{{\color[HTML]{212121} \textbf{AUDUSD}}} & \multicolumn{1}{l}{{\color[HTML]{212121} \textbf{EURUSD}}} & \multicolumn{1}{l}{{\color[HTML]{212121} \textbf{USDJPY}}} & \multicolumn{1}{l}{{\color[HTML]{212121} \textbf{EURCHF}}} & \multicolumn{1}{l}{{\color[HTML]{212121} \textbf{GBPUSD}}} & \multicolumn{1}{l}{\textbf{Average}} \\
{\color[HTML]{212121} LSTM} & {\color[HTML]{212121} ACC} & \cellcolor[HTML]{BDD881}{\color[HTML]{212121} 71,43\%} & \cellcolor[HTML]{D3DF82}{\color[HTML]{212121} 64,29\%} & \cellcolor[HTML]{D3DF82}{\color[HTML]{212121} 64,29\%} & \cellcolor[HTML]{BDD881}{\color[HTML]{212121} 71,43\%} & \cellcolor[HTML]{E9E583}{\color[HTML]{212121} 57,14\%} & \cellcolor[HTML]{CEDD82}{\color[HTML]{212121} 65,72\%} \\
{\color[HTML]{212121} LSTM+Sentiments} & {\color[HTML]{212121} ACC} & \cellcolor[HTML]{BDD881}{\color[HTML]{212121} 71,43\%} & \cellcolor[HTML]{D3DF82}{\color[HTML]{212121} 64,29\%} & \cellcolor[HTML]{BDD881}{\color[HTML]{212121} 71,43\%} & \cellcolor[HTML]{FBB379}{\color[HTML]{212121} 28,57\%} & \cellcolor[HTML]{BDD881}{\color[HTML]{212121} 71,43\%} & \cellcolor[HTML]{DCE182}{\color[HTML]{212121} 61,43\%} \\
{\color[HTML]{212121} LSTM+Sentiments+Explanations (granite)} & {\color[HTML]{212121} ACC} & \cellcolor[HTML]{BDD881}{\color[HTML]{212121} 71,43\%} & \cellcolor[HTML]{D3DF82}{\color[HTML]{212121} 64,29\%} & \cellcolor[HTML]{90CB7E}{\color[HTML]{212121} 85,71\%} & \cellcolor[HTML]{BDD881}{\color[HTML]{212121} 71,43\%} & \cellcolor[HTML]{E9E583}{\color[HTML]{212121} 57,14\%} & \cellcolor[HTML]{C1DA81}{\color[HTML]{212121} 70,00\%} \\
{\color[HTML]{212121} LSTM+Sentiments+Explanations (gpt)} & {\color[HTML]{212121} ACC} & \cellcolor[HTML]{BDD881}{\color[HTML]{212121} 71,43\%} & \cellcolor[HTML]{A6D27F}{\color[HTML]{212121} 78,57\%} & \cellcolor[HTML]{BDD881}{\color[HTML]{212121} 71,43\%} & \cellcolor[HTML]{BDD881}{\color[HTML]{212121} 71,43\%} & \cellcolor[HTML]{BDD881}{\color[HTML]{212121} 71,43\%} & \cellcolor[HTML]{B8D780}{\color[HTML]{212121} 72,86\%} \\
{\color[HTML]{212121} LSTM+Sentiments+Explanations (gpt cur)} & {\color[HTML]{212121} ACC} & \cellcolor[HTML]{BDD881}{\color[HTML]{212121} 71,43\%} & \cellcolor[HTML]{7AC57D}{\color[HTML]{212121} 92,86\%} & \cellcolor[HTML]{D3DF82}{\color[HTML]{212121} 64,29\%} & \cellcolor[HTML]{BDD881}{\color[HTML]{212121} 71,43\%} & \cellcolor[HTML]{E9E583}{\color[HTML]{212121} 57,14\%} & \cellcolor[HTML]{BDD881}{\color[HTML]{212121} 71,43\%}
\end{tabular}%
}
\end{table*}

For the prediction of currency-pair prices, data was randomly split, 60\%-training, 6\%-validation, and 34\%-testing. 
Post-model training and testing, we present predictive accuracy for all models according to three common metrics: Mean Squared Error (MSE), Mean Absolute Error (MAE), and Mean Absolute Percentage Error (MAPE). Results for all metrics are presented in table~\ref{tab:next-day-rate-prediction-accuracy} 
and shown separately for each metric, color-coded similar to the above, with higher accuracy represented by greener colors and lower accuracy indicated by redder colors. 

With the partial exclusion of the results for \texttt{USDJPY}, our results corroborate both hypotheses. We thus also show the overall average for each metric with and without the inclusion of \texttt{USDJPY}. Our first hypothesis about the inclusion of sentiment information holds across all currency pairs considering MSE, while for MAE and MAPE the same holds with only the exclusion of \texttt{USDJPY}. Similarly, our second hypothesis about the additional inclusion of sentiment-explanatory information holds among the majority of currency pairs, demonstrating the dominance of the two GPT LLMs over Granite. Here also, the \texttt{USDJPY} currency-pair shows a deviation from this pattern, for which the Granite model performs better. On average, we also see that the inclusion of k-sufficient keywords (explanations) yields further improvement in accuracy that goes beyond the enrichment of the input with only sentiment information.

\begin{table*}[ht]
\caption{Next-day price prediction for all five currency pairs.}
\vspace{-1em}
\Large
\centering
\def\arraystretch{1.5}
\resizebox{\textwidth}{!}{%
\begin{tabular}{rrrrrrrrl}
\multicolumn{1}{l}{{\color[HTML]{212121} }} & \multicolumn{1}{l}{{\color[HTML]{212121} }} & \multicolumn{1}{l}{{\color[HTML]{212121} \textbf{AUDUSD}}} & \multicolumn{1}{l}{{\color[HTML]{212121} \textbf{EURUSD}}} & \multicolumn{1}{l}{{\color[HTML]{212121} \textbf{USDJPY}}} & \multicolumn{1}{l}{{\color[HTML]{212121} \textbf{EURCHF}}} & \multicolumn{1}{l}{{\color[HTML]{212121} \textbf{GBPUSD}}} & \multicolumn{1}{l}{\textbf{Average}} & \textbf{\begin{tabular}[c]{@{}l@{}}Average\\ (-USDJPY)\end{tabular}} \\
{\color[HTML]{212121} LSTM} & {\color[HTML]{212121} MSE} & \cellcolor[HTML]{F8696B}{\color[HTML]{212121} 2,58E-04} & \cellcolor[HTML]{F8696B}{\color[HTML]{212121} 3,57E-04} & \cellcolor[HTML]{FFEB84}{\color[HTML]{212121} 2,30E+00} & \cellcolor[HTML]{F8696B}{\color[HTML]{212121} 3,93E-04} & \cellcolor[HTML]{FDBC7B}{\color[HTML]{212121} 2,77E-04} & \cellcolor[HTML]{FFEB84}{\color[HTML]{212121} 4,60E-01} & \cellcolor[HTML]{F8696B}3,21E-04 \\
{\color[HTML]{212121} LSTM+Sentiments} & {\color[HTML]{212121} MSE} & \cellcolor[HTML]{FFE082}{\color[HTML]{212121} 1,22E-04} & \cellcolor[HTML]{FFEB84}{\color[HTML]{212121} 7,82E-05} & \cellcolor[HTML]{F4E883}{\color[HTML]{212121} 2,23E+00} & \cellcolor[HTML]{FCA978}{\color[HTML]{212121} 2,19E-04} & \cellcolor[HTML]{FFEB84}{\color[HTML]{212121} 2,30E-04} & \cellcolor[HTML]{F4E883}{\color[HTML]{212121} 4,46E-01} & \cellcolor[HTML]{FDBC7B}1,62E-04 \\
{\color[HTML]{212121} LSTM+Sentiments+Explanations (granite)} & {\color[HTML]{212121} MSE} & \cellcolor[HTML]{63BE7B}{\color[HTML]{212121} 4,20E-05} & \cellcolor[HTML]{FDB97B}{\color[HTML]{212121} 1,87E-04} & \cellcolor[HTML]{63BE7B}{\color[HTML]{212121} 1,23E+00} & \cellcolor[HTML]{63BE7B}{\color[HTML]{212121} 9,33E-06} & \cellcolor[HTML]{F8696B}{\color[HTML]{212121} 3,59E-04} & \cellcolor[HTML]{63BE7B}{\color[HTML]{212121} 2,45E-01} & \cellcolor[HTML]{F4E883}1,49E-04 \\
{\color[HTML]{212121} LSTM+Sentiments+Explanations (gpt4.0)} & {\color[HTML]{212121} MSE} & \cellcolor[HTML]{F9E983}{\color[HTML]{212121} 1,06E-04} & \cellcolor[HTML]{63BE7B}{\color[HTML]{212121} 5,09E-05} & \cellcolor[HTML]{FDBC7B}{\color[HTML]{212121} 4,20E+00} & \cellcolor[HTML]{FFEB84}{\color[HTML]{212121} 3,86E-05} & \cellcolor[HTML]{63BE7B}{\color[HTML]{212121} 5,68E-05} & \cellcolor[HTML]{FDBC7B}{\color[HTML]{212121} 8,40E-01} & \cellcolor[HTML]{63BE7B}6,31E-05 \\
{\color[HTML]{212121} LSTM+Sentiments+Explanations (gpt3.5)} & {\color[HTML]{212121} MSE} & \cellcolor[HTML]{FFEB84}{\color[HTML]{212121} 1,09E-04} & \cellcolor[HTML]{96CC7D}{\color[HTML]{212121} 5,99E-05} & \cellcolor[HTML]{F8696B}{\color[HTML]{212121} 7,54E+00} & \cellcolor[HTML]{70C17B}{\color[HTML]{212121} 1,18E-05} & \cellcolor[HTML]{85C77C}{\color[HTML]{212121} 9,48E-05} & \cellcolor[HTML]{F8696B}{\color[HTML]{212121} 1,51E+00} & \cellcolor[HTML]{63BE7B}6,88E-05 \\
\multicolumn{1}{l}{{\color[HTML]{212121} }} & \multicolumn{1}{l}{{\color[HTML]{212121} }} & \multicolumn{1}{l}{{\color[HTML]{212121} }} & \multicolumn{1}{l}{{\color[HTML]{212121} }} & \multicolumn{1}{l}{{\color[HTML]{212121} }} & \multicolumn{1}{l}{{\color[HTML]{212121} }} & \multicolumn{1}{l}{{\color[HTML]{212121} }} & \multicolumn{1}{l}{{\color[HTML]{212121} }} &  \\[-2ex]
{\color[HTML]{212121} LSTM} & {\color[HTML]{212121} MAE} & \cellcolor[HTML]{F8696B}{\color[HTML]{212121} 1,47E-02} & \cellcolor[HTML]{F8696B}{\color[HTML]{212121} 1,74E-02} & \cellcolor[HTML]{B2D47F}{\color[HTML]{212121} 1,08E+00} & \cellcolor[HTML]{F8696B}{\color[HTML]{212121} 1,95E-02} & \cellcolor[HTML]{FDBA7B}{\color[HTML]{212121} 1,48E-02} & \cellcolor[HTML]{BBD780}{\color[HTML]{212121} 2,29E-01} & \cellcolor[HTML]{F8696B}1,66E-02 \\
{\color[HTML]{212121} LSTM+Sentiments} & {\color[HTML]{212121} MAE} & \cellcolor[HTML]{FFEA84}{\color[HTML]{212121} 8,97E-03} & \cellcolor[HTML]{FFEB84}{\color[HTML]{212121} 7,24E-03} & \cellcolor[HTML]{FFEB84}{\color[HTML]{212121} 1,30E+00} & \cellcolor[HTML]{FCA777}{\color[HTML]{212121} 1,30E-02} & \cellcolor[HTML]{FFEB84}{\color[HTML]{212121} 1,35E-02} & \cellcolor[HTML]{FFEB84}{\color[HTML]{212121} 2,69E-01} & \cellcolor[HTML]{FDBC7B}1,07E-02 \\
{\color[HTML]{212121} LSTM+Sentiments+Explanations (granite)} & {\color[HTML]{212121} MAE} & \cellcolor[HTML]{63BE7B}{\color[HTML]{212121} 5,48E-03} & \cellcolor[HTML]{FCAC78}{\color[HTML]{212121} 1,22E-02} & \cellcolor[HTML]{63BE7B}{\color[HTML]{212121} 8,53E-01} & \cellcolor[HTML]{63BE7B}{\color[HTML]{212121} 2,63E-03} & \cellcolor[HTML]{F8696B}{\color[HTML]{212121} 1,69E-02} & \cellcolor[HTML]{63BE7B}{\color[HTML]{212121} 1,78E-01} & \cellcolor[HTML]{F4E883}9,31E-03 \\
{\color[HTML]{212121} LSTM+Sentiments+Explanations (gpt4.0)} & {\color[HTML]{212121} MAE} & \cellcolor[HTML]{FFEB84}{\color[HTML]{212121} 8,90E-03} & \cellcolor[HTML]{63BE7B}{\color[HTML]{212121} 5,84E-03} & \cellcolor[HTML]{FDB67A}{\color[HTML]{212121} 1,77E+00} & \cellcolor[HTML]{FFEB84}{\color[HTML]{212121} 5,75E-03} & \cellcolor[HTML]{63BE7B}{\color[HTML]{212121} 5,97E-03} & \cellcolor[HTML]{FDB77A}{\color[HTML]{212121} 3,60E-01} & \cellcolor[HTML]{63BE7B}6,62E-03 \\
{\color[HTML]{212121} LSTM+Sentiments+Explanations (gpt3.5)} & {\color[HTML]{212121} MAE} & \cellcolor[HTML]{EDE582}{\color[HTML]{212121} 8,51E-03} & \cellcolor[HTML]{DCE182}{\color[HTML]{212121} 6,93E-03} & \cellcolor[HTML]{F8696B}{\color[HTML]{212121} 2,46E+00} & \cellcolor[HTML]{63BE7B}{\color[HTML]{212121} 2,62E-03} & \cellcolor[HTML]{91CB7D}{\color[HTML]{212121} 8,21E-03} & \cellcolor[HTML]{F8696B}{\color[HTML]{212121} 4,97E-01} & \cellcolor[HTML]{63BE7B}6,57E-03 \\
\multicolumn{1}{l}{{\color[HTML]{212121} }} & \multicolumn{1}{l}{{\color[HTML]{212121} }} & \multicolumn{1}{l}{{\color[HTML]{212121} }} & \multicolumn{1}{l}{{\color[HTML]{212121} }} & \multicolumn{1}{l}{{\color[HTML]{212121} }} & \multicolumn{1}{l}{{\color[HTML]{212121} }} & \multicolumn{1}{l}{{\color[HTML]{212121} }} & \multicolumn{1}{l}{{\color[HTML]{212121} }} &  \\[-2ex]
{\color[HTML]{212121} LSTM} & {\color[HTML]{212121} MAPE} & \cellcolor[HTML]{F8696B}{\color[HTML]{212121} 2,20\%} & \cellcolor[HTML]{F8696B}{\color[HTML]{212121} 1,59\%} & \cellcolor[HTML]{B0D47F}{\color[HTML]{212121} 0,81\%} & \cellcolor[HTML]{F8696B}{\color[HTML]{212121} 1,99\%} & \cellcolor[HTML]{FDB77A}{\color[HTML]{212121} 1,19\%} & \cellcolor[HTML]{F8696B}{\color[HTML]{212121} 1,56\%} & \cellcolor[HTML]{F8696B}1,74\% \\
{\color[HTML]{212121} LSTM+Sentiments} & {\color[HTML]{212121} MAPE} & \cellcolor[HTML]{FFEB84}{\color[HTML]{212121} 1,34\%} & \cellcolor[HTML]{FFEB84}{\color[HTML]{212121} 0,66\%} & \cellcolor[HTML]{FFEB84}{\color[HTML]{212121} 0,98\%} & \cellcolor[HTML]{FBA076}{\color[HTML]{212121} 1,32\%} & \cellcolor[HTML]{FFEB84}{\color[HTML]{212121} 1,08\%} & \cellcolor[HTML]{FECD7F}{\color[HTML]{212121} 1,08\%} & \cellcolor[HTML]{FDBC7B}1,10\% \\
{\color[HTML]{212121} LSTM+Sentiments+Explanations (granite)} & {\color[HTML]{212121} MAPE} & \cellcolor[HTML]{63BE7B}{\color[HTML]{212121} 0,82\%} & \cellcolor[HTML]{FCAD78}{\color[HTML]{212121} 1,11\%} & \cellcolor[HTML]{63BE7B}{\color[HTML]{212121} 0,64\%} & \cellcolor[HTML]{63BE7B}{\color[HTML]{212121} 0,27\%} & \cellcolor[HTML]{F8696B}{\color[HTML]{212121} 1,35\%} & \cellcolor[HTML]{63BE7B}{\color[HTML]{212121} 0,84\%} & \cellcolor[HTML]{F4E883}0,89\% \\
{\color[HTML]{212121} LSTM+Sentiments+Explanations (gpt4.0)} & {\color[HTML]{212121} MAPE} & \cellcolor[HTML]{FFEA84}{\color[HTML]{212121} 1,35\%} & \cellcolor[HTML]{63BE7B}{\color[HTML]{212121} 0,54\%} & \cellcolor[HTML]{FDBA7B}{\color[HTML]{212121} 1,29\%} & \cellcolor[HTML]{FFEB84}{\color[HTML]{212121} 0,38\%} & \cellcolor[HTML]{B8D67F}{\color[HTML]{212121} 0,89\%} & \cellcolor[HTML]{BBD780}{\color[HTML]{212121} 0,89\%} & \cellcolor[HTML]{BBD780}0,79\% \\
{\color[HTML]{212121} LSTM+Sentiments+Explanations (gpt3.5)} & {\color[HTML]{212121} MAPE} & \cellcolor[HTML]{EFE683}{\color[HTML]{212121} 1,29\%} & \cellcolor[HTML]{E5E382}{\color[HTML]{212121} 0,64\%} & \cellcolor[HTML]{F8696B}{\color[HTML]{212121} 1,79\%} & \cellcolor[HTML]{63BE7B}{\color[HTML]{212121} 0,27\%} & \cellcolor[HTML]{63BE7B}{\color[HTML]{212121} 0,66\%} & \cellcolor[HTML]{FFEB84}{\color[HTML]{212121} 0,93\%} & \cellcolor[HTML]{63BE7B}0,72\%
\end{tabular}%
}
\label{tab:next-day-rate-prediction-accuracy}
\end{table*}


\section{Related Work}

We relate our work to the literature in the domain of explainability for SA. 


Generative AI and transformer-based LLMs like BERT~\cite{Davlin2018} and GPT~\cite{Redford2018} have excelled in sentiment analysis among various tasks, with BERT significantly advancing general language understanding tasks~\cite{Mutinda2023}. However, their large size and weight count lead to slower training. GPT-3.5's performance further highlights LLMs' superiority in current methods~\cite{Kheiri2023}. Our research extends LLMs' utility to explain other ML models used for SA, challenging conventional post-hoc XAI tools like SHAP or LIME for explainability~\cite{Kheiri2023}. Recent studies, including one on financial texts, indicate that LLMs can enhance sentiment classification performance by 35\% and similarly improve correlations with market returns~\cite{Fatouros2023}.
While this work shows the potential benefit of utilizing LLMs for sentiment classification and in demonstrating that such sentiments retain a level of about $0.6-0.7$ correlation with market trends, our work extends this work further in two aspects. We show that the enrichment of a ML model input with not only the core sentiments associated with market prices, but also with the set of K-sufficient terms associated with each such sentiment, could improve the accuracy of such a model in predicting market trends. Subsequently, we also show that the model is fairly capable of better predicting not only the market trends with respect to individual currency-pairs, but also the actual future prices.


The work in~\cite{Fahland2024} shows LLMs' potential for explaining business process models, offering a simpler alternative to traditional XAI tools~\cite{Mavrepis2024}. LLMs, despite their complex architecture, can improve result understanding and trust through LLM-based explanations~\cite{Ding2024}. A survey by Zhao et al.~\cite{Zhao2024} highlights various LLM explainability techniques based on the training paradigms of LLMs.

Probably the closest to our work on using LLMs to explain sentiment analysis results 
is the work in~\cite{Bhattacharjee2023} where LLMs are being employed for generating counterfactual explanations. It presents a systematic approach to identify 
a minimal set of text elements that changes the classification outcomes. Following the use of LLMs for explaining SA results, we can rely on their application for the realization of the optional minimality step in our method for SA explainability. 

While the interpretability of explanations for specific users can only be verified through empirical assessment, as pursued in~\cite{Fahland2024}, LLMs can also assist in automating the fidelity check of explanations. As indicated in~\cite{Meske2022}, LLMs can contribute to the factuality evaluation of a given summary to assess its faithfulness.

Our core innovation is in the post-hoc methodological approach for leveraging an LLM to identify a sufficient set of K-terms that retains the original classification of the SA model as a novel approach, inspired by counterfactual explanations in adversarial machine-learning in which minimal sufficient sets of terms may typically be sought to reverse a model's classification~\cite{Watson2022LocalPractice}. 
Our approach leverages LLMs to infer keywords
based on their inherent generative power derived from the richness of underlying corpora used in their development, 
as opposed to employing some XAI algorithmic convergence approach as in~\cite{Carter2020}. Furthermore, we exploit these keywords to improve the accuracy of the ML model predictions of market prices.


\section{Conclusions and Future Work}

Our results support the two hypotheses. While we may have limited the scope of our empirical investigation presented to a small set of currency pairs during 3 months, our results look promising concerning both envisioned concepts, demonstrating the potential monetary benefit of leveraging recent LLM technology for the financial domain and automated trading. Further investigation should aspire to increase the size of the dataset to corroborate our findings. Our results are not only consistent with recent findings that argue for the predictive value of utilizing sentiment information for the prediction of future prices but are also showing great promise in using explainability information in a way that goes beyond its conventional intent to make model predictions more interpretable to humans. As one such concrete application, our results show the value of leveraging sentiment explainability to automatically enrich the input into predictive financial trading models, where such input promotes improving its predictive accuracy. While our second hypothesis was validated only with one technique for explanation derivation and enrichment, other realization methods may be employed, such as replacing embeddings with SHAP values as the realization of the input enrichment. In general, we argue that the presented application in the financial domain is only one example domain in which this approach could be exploited, while in general, the use of explainability could yield a beneficial input enrichment in other application domains as well.

We also foresee how recent advances in 
LLMs developed for time-series predictions present a viable alternative to using models such as LSTM employed in our work and are likely to equip companies with complete out-of-the-box tool suites that could be fairly easy to compose to automate such applications.



\begin{ack}
This project has received funding from the European Union’s Horizon research and innovation programme under grant agreement no 101092639 (FAME).
\end{ack}



\bibliography{references}

\end{document}